\documentclass{article}

\usepackage{microtype}
\usepackage{graphicx}
\usepackage{subcaption}
\usepackage{booktabs} %

\usepackage{hyperref}

\usepackage[preprint]{icml2024}

\usepackage{amsmath}
\usepackage{amssymb}
\usepackage{mathtools}
\usepackage{amsthm}

\usepackage[capitalize,noabbrev]{cleveref}

\theoremstyle{plain}

\theoremstyle{definition}

\theoremstyle{remark}

\usepackage[textsize=tiny]{todonotes}

\usepackage{enumitem}

\usepackage{tikz}
\usetikzlibrary{bayesnet}
\usetikzlibrary{arrows}
\usetikzlibrary{decorations}

\usepackage{tabularx}
\usepackage{tabulary}
\usepackage{colortbl}
\usepackage{xspace}
\usepackage{microtype}
\usepackage{xcolor}
\newcolumntype{Y}{>{\centering\arraybackslash}X}
  \newcolumntype{P}{>{\raggedleft\arraybackslash}X}
\usepackage{multirow}
\usepackage{stfloats}

\newcommand{\sft}{SFT\xspace}
\newcommand{\spiel}{SpIEL\xspace}
\newcommand{\sftrigl}{{SpIEL-AG}\xspace}
\newcommand{\sftsm}{{SpIEL-MA}\xspace}
\DeclareMathOperator{\scatter}{scatter}
\DeclareMathOperator{\argtopk}{argtopk}
\DeclareMathOperator{\reshape}{reshape}
\DeclareMathSymbol{\shortminus}{\mathbin}{AMSa}{"39}

\usepackage{siunitx}

\usepackage[noend]{algpseudocode}

\usepackage{soul}
\DeclareRobustCommand{\hlcyan}[1]{{\sethlcolor{cyan!30}\hl{#1}}}
\DeclareRobustCommand{\hlred}[1]{{\sethlcolor{red!30}\hl{#1}}}
\DeclareRobustCommand{\hlyellow}[1]{{\sethlcolor{yellow!30}\hl{#1}}}

\usepackage{amsmath,amsfonts,bm}

\def\eqref#1{equation~\ref{#1}}

\def\1{\bm{1}}

\def\vdelta{{\bm{\delta}}}
\def\vtheta{{\bm{\theta}}}

\def\vphi{{\bm{\phi}}}

\def\veta{{\bm{\eta}}}

\def\vb{{\bm{b}}}

\DeclareMathAlphabet{\mathsfit}{\encodingdefault}{\sfdefault}{m}{sl}
\SetMathAlphabet{\mathsfit}{bold}{\encodingdefault}{\sfdefault}{bx}{n}

\def\gD{{\mathcal{D}}}

\def\gL{{\mathcal{L}}}

\def\sI{{\mathbb{I}}}

\def\sR{{\mathbb{R}}}

\def\din{d_{\text{in}}}
\def\dout{d_{\text{out}}}

\newcommand{\E}{\mathbb{E}}

\DeclareMathOperator*{\argmax}{arg\,max}

\usepackage{todonotes}
\makeatletter
\newcommand*\iftodonotes{\if@todonotes@disabled\expandafter\@secondoftwo\else\expandafter\@firstoftwo\fi}
\makeatother

\definecolor{edolime}{rgb}{0.9,1,0.3}

\definecolor{alanmag}{rgb}{1.0,0,1.0}

\definecolor{ivanc}{rgb}{0.0,0.8,0.9}

\icmltitlerunning{Scaling Sparse Fine-Tuning to Large Language Models}

\begin{document}

\newcommand{\edin}{{1583}}
\newcommand{\cam}{{1209}}

\twocolumn[
\icmltitle{Scaling Sparse Fine-Tuning to Large Language Models}

\begin{icmlauthorlist}
\icmlauthor{Alan Ansell}{cam}
\icmlauthor{Ivan Vulić}{cam}
\icmlauthor{Hannah Sterz}{cam}
\icmlauthor{Anna Korhonen}{cam}
\icmlauthor{Edoardo M. Ponti}{edin,cam}
\end{icmlauthorlist}

\icmlaffiliation{cam}{University of Cambridge}
\icmlaffiliation{edin}{University of Edinburgh}

\icmlcorrespondingauthor{Alan Ansell}{\texttt{aja63@cam.ac.uk}}

\icmlkeywords{Machine Learning, ICML}

\date{}

\vskip 0.3in
]

\printAffiliationsAndNotice{}

\begin{abstract}
Large Language Models (LLMs) are difficult to fully fine-tune (e.g., with instructions or human feedback) due to their sheer number of parameters. A family of \textit{parameter}-efficient sparse fine-tuning methods have proven promising in terms of performance but their \textit{memory} requirements increase proportionally to the size of the LLMs. In this work, we scale sparse fine-tuning to state-of-the-art LLMs like LLaMA 2 7B and 13B.
We propose \spiel, a novel sparse fine-tuning method which, for a desired density level, maintains an array of parameter indices and the deltas of these parameters relative to their pretrained values. It iterates over: (a) updating the active deltas, (b) pruning indices (based on the change of magnitude of their deltas) and (c) regrowth of indices. For regrowth, we explore two criteria based on either the accumulated gradients of a few candidate parameters or their approximate momenta estimated using the efficient SM3 optimizer. 
We experiment with instruction-tuning of LLMs on standard dataset mixtures, finding that \spiel is often superior to popular parameter-efficient fine-tuning methods like LoRA (low-rank adaptation) in terms of performance and comparable in terms of run time. We additionally show that \spiel is compatible with both quantization and efficient optimizers, to facilitate scaling to ever-larger model sizes. We release the code for \spiel at \url{https://github.com/AlanAnsell/peft} and for the instruc\-tion-tuning experiments at \url{https://github.com/ducdauge/sft-llm}.
\end{abstract}

\begin{figure*}
    \centering
    \includegraphics[width=0.9\textwidth]{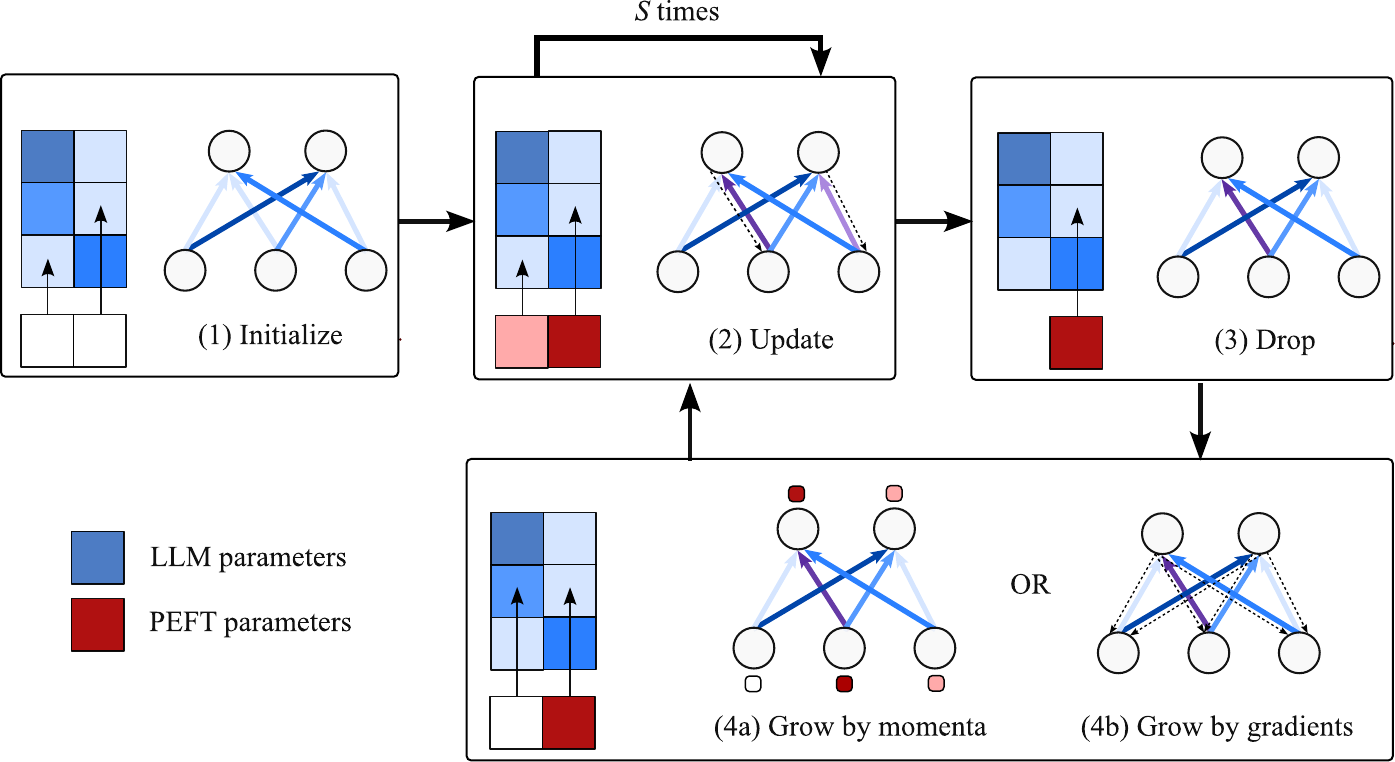}
    \caption{A visualization of the proposed Sparse Fine-Tuning (\sft) method scaled to a Large Language Model (LLM). PEFT parameters consist of indices (arrows) and corresponding deltas (red squares) with respect to LLM parameters (blue squares). After initialization (1), PEFT deltas are updated for $S$ steps (2). Next, obsolete indices are dropped (3) and new indices are grown (4) according to either accumulated gradients or approximate momenta. The algorithm then returns to the update step (2) and is repeated iteratively.}
    \label{fig:sft}
    \vskip -0.1in
\end{figure*}

\section{Introduction}
The scale of Large Language Models (LLMs), such as Falcon~\citep{falcon40b}, LLaMA 2~\citep{touvron2023llama}, and Mistral~\citep{jiang2023mistral}, is one of the keys to their state-of-the-art performance~\citep{kaplan2020scaling}. However, this scale is both a blessing and a curse as tailoring LLMs to specific applications via fine-tuning presents a formidable challenge: if performed na\"ively, this incurs the cost of updating an incredibly large set of parameters. A family of lightweight methods for LLM adaptation have been proposed to mitigate this issue, known collectively as Parameter-Efficient Fine-Tuning (PEFT). PEFT methods learn a small number of new parameters, denoted as $\vphi$, which augment the frozen LLM weights $\vtheta$ \citep{pfeiffer2023modular,lialin2023scaling}. For instance, Low-Rank Adapters \citep[LoRA;][]{hu2022lora} learn additional low-rank matrices to modify the linear layers in Transformer blocks.

PEFT methods based on unstructured sparse fine-tuning (\sft\footnote{Note the unfortunate confusion of nomenclature with \textit{supervised} fine-tuning (also frequently referred to as \sft).}),
where $\vphi$ is a sparse vector added to $\vtheta$, have recently shown promise \citep{sung2021training,guo-etal-2021-parameter,ansell-etal-2022-composable}. These offer a strong trade-off between low number of parameters and high model performance without inserting additional layers into the LLM's neural architecture, which would reduce model efficiency. In addition, multiple {\sft}s are \textit{composable} while avoiding interference \citep{ansell-etal-2022-composable}, which facilitates the integration of multiple sources of knowledge into LLMs.
Formally, \sft can be conceived of as performing joint optimization over the fixed-size set of non-zero indices of $\vphi$ and their deltas with respect to the LLM weights.
Due to the intricacies of this optimization, however, \sft has so far been severely limited by a major drawback, namely, its high memory requirements: existing methods for selecting non-zero indices include learning a mask~\citep{sung2021training}, estimating the Fisher information \citep{guo-etal-2021-parameter}, or calculating the difference between initialization and convergence \citep{ansell-etal-2022-composable} for \textit{all LLM parameters}. Hence, \sft is not currently suitable for adapting LLM at large scales.

The main goal of this work is to overcome these challenges by devising memory-efficient methods to update Large Language Models (LLMs) sparsely, while maintaining performance benefits, that is, retaining the same performance of full-model fine-tuning or even surpassing it. Specifically, we wish for the memory use during training (beyond that required to store the pretrained model weights) to scale linearly with the number of \sft parameters $\mathcal{O}(d_\vphi)$ rather than LLM parameters $\mathcal{O}(d_\vtheta)$. To achieve this, we introduce SpIEL (Sparse Iterative Efficient Learning), an iterative paradigm for \sft that alternates between updating deltas of active indices, deleting obsolete indices, and growing new ones~\citep{evci-etal-2020-rigging}. %
Deletion is determined by change of magnitude between training steps whereas growth by (transiently calculated) gradients. The growth criterion is inspired by \citet{evci-etal-2020-rigging}, which we further improve upon by \textit{efficiently accumulating} gradients to reduce their variance. 
Moreover, while our algorithm (\sftrigl) is reminiscent of sparse training \citep{evci-etal-2020-rigging}, the scattering of the \sft onto the base LLM effectively yields a dense model. This entirely side-steps the problem of the `hardware lottery' due to cumbersome sparse tensor operations \citep{hooker2020hardware}. We provide a visual overview of our algorithms in Figure \ref{fig:sft}.

When extreme memory efficiency is required, we show how \spiel can be combined with efficient optimizers such as SM3, where the momenta of parameter matrices are approximated by row-wise and column-wise summary metrics \citep{anil2019memory}. In these settings, gradients become unreliable, as their variance increases in single-example mini-batches and it is costly memory-wise to accumulate even a subset of them. Hence, we propose that approximate momenta can additionally substitute gradients as a criterion for growth, yielding the \sftsm model variant.

We compare our \spiel variants with the state-of-the-art PEFT methods LoRA \citep{hu2022lora} and (IA)$^3$~\citep{ia3}, as well as with full fine-tuning, starting from LLaMA 2~\citep{touvron2023llama} as a base model. We instruction-tune them on multi-task data such as Flan v2 \citep{longpre2023flan}, data generated by proprietary models such as GPT4-Alpaca \citep{peng2023instruction}, or a mixture of both with Tülu v2 \citep{ivison2023camels}. We extensively evaluate the resulting models on standard benchmarks for factuality \citep[MMLU;][]{hendrycks2021measuring}, reasoning \citep[GSM and BBH;][]{cobbe2021gsm8k,suzgun2022challenging}, multilinguality \citep[TyDiQA;][]{clark2020tydiqa}, and coding \citep[HumanEval;][]{chen2021evaluating}.

The main results reveal that \spiel outperforms the PEFT and full fine-tuning baselines on most tasks and configurations we test, both with and without 4-bit LLM quantization during fine-tuning \citep{dettmers2023qlora}. In combination with the \sftsm variant, this allows for scaling fine-tuning to very large LLMs with a modest memory footprint.

\section{Background and Related Work}

\subsection{Parameter-Efficient and Memory-Efficient Fine-Tuning}

Parameter-efficient fine-tuning (PEFT) methods have generally been defined as those which fine-tune a small number of parameters relative to the total size of the pretrained model. The possible benefits of using a PEFT method as opposed to full model fine-tuning include: (1) reduced GPU memory usage during training; (2) faster training; (3) faster saving and loading of the fine-tuning with less permanent storage required as the ``frozen'' original weights of the large underlying model are shared across multiple tasks and applications; (4) the ``composability'' property of some PEFT methods, which allows modules to be combined with less interference than with full fine-tuning \citep{pfeiffer-etal-2020-mad, ansell-etal-2022-composable}; and (5) less tendency to overfit due to the reduced capacity of PEFT with respect to the full model.

Of these, (1) is perhaps the most critical in the era of Large Language Models whose GPU memory requirement for full fine-tuning is beyond the reach of researchers and developers without multiple high-end GPUs. Nonetheless, parameter-efficiency alone does not guarantee a reduction in GPU memory usage, though it almost certainly implies that less space is required to save the fine-tuning in permanent memory. It is thus important to draw a distinction between (i) efficiency in number of fine-tuned parameters versus (ii) the peak GPU memory usage during fine-tuning: we refer to the former as \textit{parameter efficiency} and the latter as \textit{memory efficiency}.

\subsection{LoRA}
\label{ssec:lora}
As a solution,
Low-Rank Adaptation \citep[LoRA;][]{hu2022lora} is of the best-performing and most popular PEFT techniques to date. In brief, it conceptually fine-tunes only a low-rank subspace of each weight matrix. In practice, LoRA is typically applied only to the linear modules of a Transformer, and is implemented as follows:
\begin{align}
    \bm{y} = W\bm{x} + \frac{\alpha}{r} BA\bm{x},
\end{align}
where $\bm{x} \in \sR^{d_{\text{in}}}$, $\bm{y} \in \sR^{d_{\text{out}}}$, $W \in \sR^{\dout \times \din}$, $A \in \sR^{r \times \din}$, $B \in \sR^{\dout \times r}$, the subspace rank $r \ll \din, \dout$, and $\alpha$ is a hyperparameter. Like bottleneck adapters \citep{houlsby-etal-2019-parameter}, LoRA employs a successive down- and up-projection, but the fact that it places them in parallel rather than in series with the full-rank matrix multiplication without a non-linearity enables LoRA adapters to be ``merged'' at inference time into their $W$ matrix as follows:
\begin{align}
    \bm{y} = {W}^\prime\bm{x} = (W + \frac{\alpha}{r} BA)\bm{x}.
\end{align}
This allows LoRA-adapted models to achieve inference speed equal to the underlying LLM.

With the recent development of efficient quantization methods suitable for LLMs, LoRA has emerged as the \textit{de facto} standard PEFT method for LLM instruction tuning \citep{alpaca-lora, dettmers2023qlora}. This is also why we provide its short self-contained description here and treat it as our principal baseline, while we conduct additional comparisons to (IA)$^3$~\cite{ia3}, another established PEFT method.

\subsection{Sparse Fine-Tuning}
\label{ss:sft-background}
A sparse fine-tuning ${f}^\prime$ of a neural function $f$ entails the addition of a sparse ``difference'' or ``delta'' vector $\vdelta \in \sR^{d_\vtheta}$ to its parameters $\vtheta \in \sR^{d_\vtheta}$, i.e. ${f}^\prime(\ ;\ \vtheta) =f(\ ;\ \vtheta + \vdelta)$. A delta vector $\vdelta$ with $d_\vphi$ non-zero values can be expressed in terms of a vector of unique indices $\veta \in \{1, 2, ..., d_\vtheta\}^{d_\vphi}$ and their corresponding values $\vphi \in \sR^{d_\vphi}$. Typically, the \sft density $\frac{d_\vphi}{d_\vtheta}$ is a user-definable hyperparameter. To illustrate the difference between LoRA and \sft visually, Figure \ref{fig:peft} shows how they adapt a Transformer block.

In its most general form, sparse fine-tuning can be interpreted as performing joint optimization over $\veta$ and $\vphi$:
\begin{equation}
    \veta^\star, \vphi^\star = \argmax_{\veta, \vphi} \log p(\mathcal{D} \mid \vtheta, \veta, \vphi).
\end{equation}
A specialized optimization approach is required for $\veta$ since it is a discrete latent variable and cannot be directly optimized through stochastic gradient descent. Approaches proposed in previous works include: \textit{DiffPruning} \citep{guo-etal-2021-parameter}, which applies a continuous relaxation of a binary mask to $\vdelta$ during fine-tuning which is sparsified with a regularization term, and takes $\veta$ to be the $d_\vphi$ indices with mask values closest to 1 at the end; \textit{FISH Mask} \citep{sung2021training}, where $\veta$ is fixed at the beginning of training to be the indices of the $d_\vphi$ weights with highest observed Fisher information; and \textit{Lottery-Ticket \sft} \citep[LT-\sft;][]{ansell-etal-2022-composable}, where $\veta$ is fixed to be the indices of the $d_\vphi$ weights which change the most during an initial round of full fine-tuning. These methods share a common drawback, namely that the amount of memory they use during training in addition to that required to store the pretrained model weights is proportional to $d_\vtheta$, the total number of model parameters. As discussed above, this makes these methods prohibitively expensive in many LLM fine-tuning scenarios, especially with very large models; we thus seek a method whose memory overhead is instead proportional to $d_\vphi$, the number of parameters which are actually modified during fine-tuning.

Finally, there exists a separate but related literature on \textit{pre-training} sparse neural networks, for which we refer the reader to \citet{hoefler2021sparsity} for a detailed overview. This work owes most to \citet{evci-etal-2020-rigging}, whose dropping and growth paradigm we extend to sparse fine-tuning in \S\ref{ssec:sft-rigl}. Contrary to them, however, \spiel results in a \textit{dense} model, which side-steps the problem that available hardware is not well suited to sparse tensor operations \citep{hooker2020hardware}. Moreover, we introduce novel and enhanced growth criteria in \cref{ssec:sft-rigl} and \cref{ssec:sft-sm3}.

\subsection{Quantized PEFT}
A significant recent advance in efficient methods for NLP has been the development of quantization methods which are suitable for LLMs and incur minimal degradation in performance. \citet{dettmers-etal-2022-llmint8} and later \citet{dettmers2023qlora} proposed 8-bit and 4-bit quantization techniques for LLM parameter tensors that yield close to full-precision performance during inference or parameter-efficient fine-tuning. The qLoRA fine-tuning method of \citet{dettmers2023qlora} reduces memory usage by applying 4-bit quantization to most of the pretrained LLM parameters while storing the small number of additional trainable LoRA parameters in full precision and dequantizing the pretrained weights only when required. We show that sparse fine-tuning is also amenable to quantization of the pretrained weights. We use $\textsc{quant}(\cdot)$ and $\textsc{dequant}(\cdot)$ to denote 4-bit NormalFloat quantization and dequantization respectively with double quantization \citep{dettmers2023qlora}.

\section{Method}

\subsection{Efficient \sft with Fixed $\veta$}
In practice, the weights $\vtheta$ of a neural function are partitioned into a sequence of $n_p$ ``parameter tensors.'' To simplify indexing, we think in terms of the flattened versions of these tensors and refer to them as ``parameter subvectors'', denoted as $\{\vtheta^{(1)} \in \sR^{d_{\vtheta^{(1)}}}, \vtheta^{(2)} \in \sR^{d_{\vtheta^{(2)}}}, ..., \vtheta^{(n_p)} \in \sR^{d_{\vtheta^{(n_p)}}}\}$. Similarly, we denote the sections of $\veta$ and $\vphi$ corresponding to parameter subvector $\vtheta^{(i)}$ as $\veta^{(i)}$ and $\vphi^{(i)}$ respectively. We observe that, for a fixed $\veta$ and assuming $\max_i d_{\vtheta^{(i)}} < d_\vphi$, it is possible to perform sparse fine-tuning with the desired $\mathcal{O}(d_\vphi)$ memory overhead by scatter-adding each $\vphi^{(i)}$ into its corresponding $\vtheta^{(i)}$ (which stays frozen) before it is used:
\begin{align}
    \vtheta'^{(i)} = \vtheta^{(i)} + \scatter(\veta^{(i)}, \vphi^{(i)}, d_{\vtheta^{(i)}}),
\end{align}
where $\scatter(\veta, \vphi, d) \in \sR^d$ such that
\begin{align}
    [\scatter(\veta, \vphi, d)]_i = \sum_{j=1}^{d_\vphi} \sI_{\veta_j = i}\ \vphi_j.
\end{align}
The simplest way of calculating the gradient of $\vphi^{(i)}$ during the backward pass of backpropagation is by gathering the relevant indices of the gradient of $\vtheta^{\prime(i)}$:
\begin{align} \label{eq:sft_grad}
    \frac{\partial \gL}{\partial \phi^{(i)}_j} = \frac{\partial \gL}{\partial \theta^{\prime(i)}_{\eta_j}}.
\end{align}
While \cref{eq:sft_grad} requires the computation of the dense gradient of each $\vtheta^{(i)}$, this can be disposed of as soon as the required values are gathered from it. Because $d_{\vtheta^{(i)}} \ll d_\vtheta$, this does not add significantly to the peak memory usage. Furthermore, we show in Appendix \ref{app:sft-backward} that for a linear layer, it is possible in principle to calculate the gradient of $\vphi^{(i)}$ without needing to compute the full gradient of $\vtheta^{(i)}$, which we will explore in future work.

This implementation of sparse fine-tuning stands in contrast to previous approaches \citep[e.g.][]{sung2021training,ansell-etal-2022-composable} which, instead of vectors $\veta$ and $\vphi$, maintain a binary mask $\vb^{(i)} \in \{0, 1\}^{d_{\vtheta}^{(i)}}$ which is applied to the gradient of $\vtheta^{(i)}$ after it is calculated. Since $\vb$ and the gradient of $\vtheta$ both have size proportional to $d_\vtheta$, the memory overhead of this implementation is $\mathcal{O}(d_\vtheta)$. Although the above implementation would enable the fixed-$\veta$ stage of FISH Mask or LT-\sft to be performed with acceptable memory overhead, both methods incur $\mathcal{O}(d_\vtheta)$ memory cost when selecting $\veta$.

\subsection{\sftrigl: Accumulated Gradient \spiel} \label{ssec:sft-rigl}
Building on \citet{evci-etal-2020-rigging}'s ``Rigging the Lottery'' (RigL) method for pretraining sparse neural networks, we propose \sftrigl. \sftrigl maintains a fixed-size $\veta^{(i)}$ and $\vphi^{(i)}$ for each parameter subvector $\vtheta^{(i)}$, but unlike the methods discussed in the previous section, it allows $\veta^{(i)}$ to change dynamically during training. $\vphi^{(i)}$ is initialized\footnote{The total number of tunable parameters $d_\vphi$ is a hyper-parameter, and $d_{\vphi^{(i)}}$ is set such that the proportion of tunable parameters $\frac{d_{\vphi^{(i)}}}{d_{\vtheta^{(i)}}}$ is the same for each $i$-th parameter subvector.} as $[0]^{d_{\vphi^{(i)}}}$ and $\veta^{(i)}$ is initialized as a random subset of $\{1, 2, ..., d_{\vtheta^{(i)}}\}$. Every $S$ training steps, $\veta^{(i)}$ is updated by freezing some of the currently trainable weights after resetting them to their pretrained values (\textit{``dropping''}), while unfreezing some of the currently frozen weights (\textit{``growth''}). A family of possible \spiel methods arises from the choice of criteria for dropping and growth. For \sftrigl, we use the following criteria:
\begin{itemize}[leftmargin=*]
\setlength\itemsep{-0mm}
    \item \hlred{\textsc{Drop:}} the $k(i, t)$ weights in $\veta^{(i)}$ which have changed the least from their pretrained values, i.e. $\veta_\text{drop}^{(i)} = \argtopk(i, t)\ -|\vphi^{(i)}|$, where $k(i, t)$ is a schedule defining the number of weights in parameter subvector $i$ to replace at step $t$.
    \item \hlcyan{\textsc{Grow:}} the $k(i, t)$ weights in $\vtheta^{(i)}$ with largest estimated ``long-run'' gradient magnitudes, i.e. $\veta_\text{grow}^{(i)} = \argtopk(i, t)\ \big|\hat{\E}_{\bm{x} \sim \gD}[\nabla_{\vtheta^{(i)}} \gL(\bm{x};\ \vtheta^\prime)]\big|$.
\end{itemize}
The gradients required for growth selection are estimated over the $\gamma$ training steps before $\veta$ is updated, which we refer to as the \textit{gradient estimation phase}. Here we diverge from \citet{evci-etal-2020-rigging}, who select the weights to grow on the basis of gradients from a single instantaneous minibatch.\footnote{In a memory-constrained setting where it may be possible to process just a single training example at a time, the high level of variance in the gradients from one minibatch may harm the selection of weights to grow.} It is not possible to maintain an estimate of the full gradient of dimension $d_{\vtheta^{(i)}}$ without exceeding our memory budget. Therefore, we restrict ourselves to maintaining such an estimate for just $K_i$ ``growth candidates.'' These are selected as the weights in $\vtheta^{(i)}$ with top $K_i$ gradient magnitudes during the first batch of the gradient estimation phase. The long-run gradients of the growth candidates are estimated by averaging their gradients over all batches in the gradient estimation phase:
\begin{align}
    \hat{g}_j^{(i)}(t) = \frac{1}{\gamma} \sum_{s=t-\gamma+1}^t \frac{\partial}{\partial \theta_j^{(i)}} \gL(\bm{x}_s;\ \vtheta^\prime).
\end{align}
Note that although it is necessary to calculate the dense gradient $\nabla_{\vtheta^{(i)}} \gL(\bm{x};\ \vtheta^\prime)$ at each step of the gradient estimation phase, we never need to \textit{store} it since we can immediately gather the $K_i$ values we need from it. This can be implemented with a backward hook in PyTorch, for instance. In our experiments, we set $K_i = d_{\vphi^{(i)}}$.

There is some additional housekeeping to do after updating $\veta$. We must reset $\vphi^{(i)}$ to zero at indices $\textsc{drop}(\veta^{(i)})$. We also need to set the optimizer buffers for the newly grown weights $\textsc{grow}(\veta^{(i)})$ to appropriate values. In our experiments, we use the Adam optimizer \citep{kingma-ba-2015-adam}, which tracks the exponentially smoothed mean of the first and second momenta of the parameter gradients. Conveniently, the gradient estimation phase already produces an estimate of the first momentum of the newly grown weights, and we extend it to estimate the second as well so that we can use these values to seed the optimizer momenta. A minor complication here is that Adam multiplies the $i$-th momentum by a factor of $\frac{1}{1 - \beta_i^t}$ before use to correct for its bias toward zero. Since in \spiel, different weights are ``initialized'' at different times, we track the age of each weight (i.e. the number of steps since it was last grown) individually, and use it to calculate the appropriate bias correction.

For simplicity, we set the update rate schedule $k(i, t)$ to decrease linearly to 0 over the course of training, as follows:
\begin{align}
    k(i, t) = \begin{cases}
        d_{\vphi^{(i)}} &\text{if } t = \gamma, \\
        \frac{\xi (T - t)}{T} d_{\vphi^{(i)}} &\text{otherwise,}
    \end{cases}
\end{align}
where $T$ is the total number of training steps and $\xi$ is a hyperparameter denoting the peak replacement rate. Note that during the first $\veta$ update at step $\gamma$, we replace all the weights, since the indices in $\veta$ are randomly initialized; there is no reason to believe that they are relevant to the task.

We provide high-level pseudocode for \sftrigl in Algorithm \ref{alg:sft-rigl}. Note that some details such as the selection of candidates and optimizer momenta seeding are omitted for conciseness.

\algnewcommand{\LineComment}[1]{\State \(\triangleright\) #1}

\begin{algorithm}[t]
\caption{Accumulated Gradient \spiel}\label{alg:sft-rigl}
\begin{algorithmic}[1]
\Procedure{\sftrigl}{$\vtheta, \mathcal{D}, \gamma, k$}
\State $\vphi \gets [0]^{d_\vphi}$ \Comment{Initialize \sft values}
\State $\vphi_0 \gets \vphi$
\State $\veta \sim_{d_\vphi} [1 . . d_\vtheta]$ \Comment{Initialize \sft indices}
\For{$t \text{ in } 1 . . T$}
        \State $\mathbf{x}_t \sim \mathcal{D}$
        \State $\vtheta^\prime \gets \text{scatter-add}(\vphi, \veta, \vtheta)$
        \State $\hlyellow{\textsc{update}}(\vphi, \nabla_\vphi \, \gL(f_{\vtheta^\prime}(\mathbf{x}_t)))$
        \If{$\gamma \, | \, t$}
        \State $\bm{d} \gets \text{top-}k(t) (- |\vphi - \vphi_0|)$
        \State $\vphi, \veta = \hlred{\textsc{drop}}(\vphi, \veta, \bm{d})$
        \State $\bm{g} \gets
            \text{top-}k(t) [\sum_{i=t-\gamma+1}^t \nabla_\vtheta \, \gL(f_{\vtheta^\prime}(\mathbf{x}_i))]$
        \State $\vphi, \veta = \hlcyan{\textsc{grow}}(\vphi, \veta, \bm{g})$
        \State $\vphi_0 \gets \vphi$
    \EndIf
\EndFor
\State \textbf{return} $\vphi$, $\veta$
\EndProcedure
\end{algorithmic}
\end{algorithm}

\subsection{\sftsm: Momentum-Approximation \spiel} \label{ssec:sft-sm3}
The \sftrigl method prioritizes making a high-quality selection of weights to grow when updating $\veta$, but this comes at the cost of the extra memory required during the gradient estimation phase to store the indices of the growth candidates and their estimated momenta. We propose an alternative algorithm for the most memory-constrained scenarios, which we call \sftsm, employing the SM3 memory-efficient adaptive optimizer of \citet{anil2019memory}. For a two-dimensional parameter tensor $\Theta$ of size $r \times c$, the SM3 optimizer maintains buffers $\bm{r} \in \sR^r$ and $\bm{c} \in \sR^c$, which contain running sums of the maximum squared gradients over the columns and rows of $\Theta$, respectively. We can obtain a (low-quality but cheap) estimate of the absolute value of the momentum $m_{ij}$ for each element $\theta_{ij}$ of $\Theta$ by taking the elementwise fourth root of the outer product of $\bm{r}$ and $\bm{c}$:
\begin{align}
    \hat{|m|}_{ij} = \sqrt[4]{r_{i} c_{j}}.
\end{align}
\sftsm uses this estimate to rank weights for growth, and otherwise it is the same as \sftrigl. Since the SM3 optimizer is very memory efficient, storing only $r + c$ values in its buffers for an $r \times c$ parameter tensor compared to Adam's $2rc$, and \sftsm uses no additional persistent memory to track statistics to inform $\veta$ updates, significantly less memory in total is required than for \sftrigl. %
Incidentally, we remark that the growth criterion of \sftsm assumes \textbf{locality}, i.e., that the importance of a parameter is correlated to those in the same row and column. This is reminiscent of the locality-driven synaptic growth in human brains, which results in dense hubs but a globally sparse anatomy \cite{betzel2017modular,hoefler2021sparsity}.
We provide high-level pseudocode for \sftsm in Algorithm \ref{alg:sft-sm3}.

\begin{algorithm}[th]
\caption{Momentum-Approximation \spiel}\label{alg:sft-sm3}
\begin{algorithmic}[1]
\Procedure{\sftsm}{$\vtheta^{(1)}, .., \vtheta^{(n_p)}, \mathcal{D}, \gamma, k$}
\For{$i \text{ in } 1 .. P$}
    \State $\vphi^{(i)} \gets [0]^{d_{\vphi^{(i)}}}$ %
    \State $\vphi_0^{(i)} \gets \vphi^{(i)}$
    \State $\veta^{(i)} \sim_{d_{\vphi^{(i)}}} [1 .. d_{\vtheta^{(i)}}]$ %
    \State $\mathbf{r}^{(i)} = [0]^{h^{(i)}}$ \Comment{Initialize row accumulator}
    \State $\mathbf{c}^{(i)} = [0]^{w^{(i)}}$ \Comment{Initialize column accumulator}
\EndFor
\For{$t \text{ in } 1 .. T$}
    \State $\mathbf{x} \sim \mathcal{D}$
    \State $\vtheta^\prime \gets \text{concat}_i\ \text{scatter-add}(\vphi^{(i)}, \veta^{(i)}, \vtheta^{(i)})$
    \For{$i \text{ in } 1 .. n_p$}
        \State $\hlyellow{\textsc{update-sm3}}[\vphi^{(i)}, \veta^{(i)}, \mathbf{r}^{(i)}, \mathbf{c}^{(i)},$
        \State $ \qquad \qquad \qquad \nabla_{\vphi^{(i)}} \, \gL(f_{\vtheta^\prime}(\mathbf{x}))]$
        \If{$\gamma \, | \, t$}
            \State $\bm{d}^{(i)} \gets \text{top-}k(i, t) (- |\vphi^{(i)} - \vphi_0^{(i)}|)$
            \State $\vphi^{(i)}, \veta^{(i)} = \hlred{\textsc{drop}}(\vphi^{(i)}, \veta^{(i)}, \bm{d}^{(i)})$
            \State $\bm{g}^{(i)} \gets
                \text{top-}k(i, t) [\mathbf{r}^{(i)} \otimes \mathbf{c}^{(i)}]$
            \State $\vphi^{(i)}, \veta^{(i)} = \hlcyan{\textsc{grow}}(\vphi^{(i)}, \veta^{(i)}, \bm{g}^{(i)})$
            \State $\vphi_0^{(i)} \gets \vphi^{(i)}$
        \EndIf
    \EndFor
\EndFor
\State \textbf{return} $\vphi^{(1)}, .., \vphi^{(n_p)}$, $\veta^{(1)}, .., \veta^{(n_p)}$
\EndProcedure
\end{algorithmic}
\end{algorithm}

\textbf{Regularizing \spiel}
Similar to LoRA, which is regularized by its dropout, \spiel is also likely to overfit as the model diverges from its pretrained state. We therefore regularize \spiel by applying L2 regularization to the $\vphi$ parameters in the form of weight decay with strength $\lambda$.

\textbf{Quantized \spiel}
As another contribution, we extend the proposed \spiel techniques to quantized LLMs ({``q\spiel''}). Consider parameter matrix $W_\text{PT}^{(i)} \in \sR^{h \times o}$ in its pre-trained data type (e.g., FP32). Instead of storing $W_\text{PT}^{(i)}$ itself on GPU, we store its quantized version $W_\text{NF4}^{(i)} = \textsc{quant}(W_\text{PT}^{(i)})$. During the forward pass of the linear module, q\spiel computes the following:
\begin{align}
    Y = X(\textsc{dequant}(W_\text{PT}^{(i)}) + \Delta W^{(i)}),
\end{align}
where $X \in \sR^{b \times h}$ is the input to the linear module, $Y \in \sR^{b \times o}$ is the output, and
\begin{align}
    \Delta W^{(i)} = \reshape(\scatter(\veta^{(i)}, \vphi^{(i)}, ho), [h,  o]). \notag
\end{align}
This is equivalent to the behavior of a linear module in ordinary, non-quantized \spiel except that the pretrained parameter matrix gets quantized at the beginning of training and temporarily dequantized each time it is used.

\section{Experimental Setup}
\label{s:experimental}
\subsection{Training and Evaluation Data}

To demonstrate the effectiveness of \spiel, we %
loosely base our experimental setup on that of~\citet{wang2023far}, who compare different data mixtures. In particular, we instruction-tune LLMs on (i) \citet{wang2023far}'s 50K sub-sample of Flan v2 \citep{longpre2023flan}, a dataset collecting manually annotated examples for multiple tasks augmented with instructions; (ii) GPT4-Alpaca \citep{peng2023instruction}, a dataset of 50K outputs generated by \texttt{davinci-003} and GPT-4 \citep{openai2023gpt4} prompted with inputs from Alpaca \citep{alpaca}; or (iii) the Tülu v2 mixture \citep{ivison2023camels}, consisting of 326K examples from multiple instruction following datasets.%

Following \citet{wang2023far} and \citet{ivison2023camels}, we evaluate instruction-tuned LLMs on the following benchmarks to capture a range of abilities: Massively Multitask Language Understanding \citep[MMLU;][]{hendrycks2021measuring}, Grade School Math \citep[GSM;][]{cobbe2021gsm8k}, BIG-Bench Hard \citep[BBH;][]{suzgun2022challenging}, Typologically Diverse Question Answering \citep[TyDiQA;][]{clark2020tydiqa} and HumanEval \citep{chen2021evaluating}. We report combinations of training data mixtures and downstream evaluation benchmarks where \citet{wang2023far} reports the highest gains.
See Appendix \ref{app:eval_setup} for full details of our evaluation setup.

\subsection{Models and Baselines}
As LLMs to fine-tune, we choose state-of-the-art LLaMA 2 \citep{touvron2023llama} at both 7b and 13b parameter scales. We report the performance of the unmodified ``vanilla'' models as a baseline for a series of PEFT methods. Specifically, we compare \spiel with LoRA \citep[see \S\ref{ssec:lora}]{hu2022lora}, as it offers the best performance--efficiency trade-off \citep{pfeiffer2023modular} and is arguably the most widespread pre-existing PEFT method \citep{dettmers2023qlora}, as well as (IA)$^3$ \citep{ia3}. %
We use the LoRA and (IA)$^3$ implementations in the \texttt{peft} library \citep{peft}. For the T\"ulu v2 fine-tuning mixture, we include the fully fine-tuned LLaMA 2 models of \citet{ivison2023camels} as a baseline (``FullFT''), which we would expect to provide a soft upper bound on the performance of the PEFT methods.\footnote{These follow a similar hyper-parameter setup except for a much higher maximum context length of 8,192.} At 7b scale, we also perform our own full fine-tuning on the Flan v2 and GPT4-Alpaca splits. We did not perform these experiments at 13b scale due to the computational expense.

Note that the pre-existing \sft methods (see \S\ref{ss:sft-background}) such as DiffPruning \citep{sung2021training}, FISH Mask \citep{guo-etal-2021-parameter}, and LT-\sft \citep{ansell-etal-2022-composable}
are not viable as baselines for LLM fine-tuning as their memory complexity scales as $\mathcal{O}(d_\vtheta)$, similar to full fine-tuning.
Full details of the training setup and hyperparameters can be found in Appendix \ref{app:training_details}.

\begin{table*}[t]
    \centering
    \caption{Performance of PEFT methods on a range of evaluation tasks for various instruction tuning datasets. $^\dagger$ indicates results obtained using the models of \citet{ivison2023camels}. T\"{u}lu v2 $\oplus$ GPT4-Alpaca refers to a setting where we compose the respective PEFTs.}
    \label{tab:main-results}
    \def\arraystretch{0.87}
    \vskip 0.12in
    \footnotesize
    \resizebox{\textwidth}{!}{
   \begin{tabular}{clccccccccc}
	\toprule
	 & & \multicolumn{2}{c}{Flan v2} & \multicolumn{1}{c}{GPT4-Alpaca} & \multicolumn{5}{c}{T\"{u}lu v2} & \multirow{2}{*}{\shortstack{T\"{u}lu v2 $\oplus$ \\  GPT4-Alpaca}}\\
 &&&&& \\
	\cmidrule(lr){3-4} \cmidrule(lr){5-5} \cmidrule(lr){6-11}
	 \multicolumn{2}{c}{\textbf{Model / Method}} & \textbf{MMLU} & \textbf{TyDiQA} & \textbf{HumanEval} & \textbf{MMLU} & \textbf{GSM} & \textbf{BBH} & \textbf{TyDiQA} & \textbf{HumanEval} & \textbf{HumanEval}\\
	\cmidrule(lr){1-2} \cmidrule(lr){3-4} \cmidrule(lr){5-5} \cmidrule(lr){6-10} \cmidrule(lr){11-11}
	\multirow{6}{*}{\rotatebox[origin=c]{90}{\textbf{\small Llama2-7b}}} & Original & 45.8 & 50.9 & 13.5 & 45.8 & 13.0 & 40.6 & 50.9 & 13.5 & -\\
	 & FullFT & 50.5 & 55.5 & 15.2 & \phantom{$^\dagger$}51.3$^\dagger$ & \phantom{$^\dagger$}34.5$^\dagger$ & \phantom{$^\dagger$}45.3$^\dagger$ & \phantom{$^\dagger$}56.5$^\dagger$ & \phantom{$^\dagger$}22.6$^\dagger$ & -\\
	\cmidrule(lr){2-2} \cmidrule(lr){3-4} \cmidrule(lr){5-5} \cmidrule(lr){6-10} \cmidrule(lr){11-11}
	 & (IA)$^3$ & 46.7 & 51.6 & 14.7 & 47.8 & 17.0 & 40.6 & 52.2 & 15.5 & 16.6 \\
	 & LoRA & 49.3 & 55.3 & 15.7 & 51.3 & 22.0 & 43.8 & 58.0 & 15.5 & 16.1 \\
	 & \sftrigl & \textbf{50.7} & \textbf{56.2} & 15.6 & \textbf{51.5} & \textbf{23.0} & \textbf{44.8} & \textbf{59.4} & \textbf{17.1} & \textbf{18.8} \\
	 & \sftsm & 48.8 & 55.8 & \textbf{16.2} & 49.5 & 16.5 & 44.7 & 56.7 & 15.3 & 17.0 \\
	\cmidrule(lr){1-2} \cmidrule(lr){3-4} \cmidrule(lr){5-5} \cmidrule(lr){6-10} \cmidrule(lr){11-11}
	\multirow{6}{*}{\rotatebox[origin=c]{90}{\textbf{\small Llama2-13b}}} & Original & 55.3 & 60.3 & 17.8 & 55.3 & 23.0 & 47.8 & 60.3 & 17.8 & -\\
	 & FullFT & - & - & - & \phantom{$^\dagger$}56.7$^\dagger$ & \phantom{$^\dagger$}50.0$^\dagger$ & \phantom{$^\dagger$}51.9$^\dagger$ & \phantom{$^\dagger$}62.9$^\dagger$ & \phantom{$^\dagger$}26.7$^\dagger$ & - \\
	\cmidrule(lr){2-2} \cmidrule(lr){3-4} \cmidrule(lr){5-5} \cmidrule(lr){6-10} \cmidrule(lr){11-11}
	 & (IA)$^3$ & 55.1 & 60.1 & 18.5 & 55.6 & 27.5 & 50.6 & 62.8 & 18.1 & 18.4 \\
	 & LoRA & \textbf{55.8} & 61.4 & 19.8 & \textbf{56.2} & 29.0 & \textbf{54.6} & \textbf{63.9} & 19.5 & 22.4 \\
	 & \sftrigl & \textbf{55.8} & \textbf{62.5} & \textbf{20.0} & 55.9 & \textbf{31.5} & 52.8 & 63.5 & \textbf{20.3} & \textbf{22.7} \\
	 & \sftsm & 55.5 & \textbf{62.5} & 19.9 & 55.9 & 29.0 & 51.2 & 62.8 & 20.2 & 20.8 \\
	\bottomrule
\end{tabular}
    }
    \vskip -0.1in
\end{table*}

\begin{table}[t]
    \centering
    \caption{Performance of quantized PEFT methods on a range of evaluation tasks for various instruction tuning datasets.}
    \label{tab:quantized-results}
    \vskip 0.12in
    \def\arraystretch{0.87}
    \footnotesize
    \begin{tabular}{clccc}
	\toprule
	 & & \multicolumn{2}{c}{Flan v2} & \multicolumn{1}{c}{GPT4-Alpaca}\\
	\cmidrule(lr){3-4} \cmidrule(lr){5-5}
	 \multicolumn{2}{c}{\textbf{Model / Method}} & \textbf{MMLU} & \textbf{TyDiQA} & \textbf{HumanEval}\\
	\cmidrule(lr){1-2} \cmidrule(lr){3-4} \cmidrule(lr){5-5}
	\multirow{4}{*}{\rotatebox[origin=c]{90}{\textbf{Ll2-7b}}} & Original (4-bit) & 44.8 & 50.2 & 11.0 \\
	 & qLoRA & 47.7 & 54.3 & 13.3 \\
	 & q\sftrigl & \textbf{48.8} & \textbf{54.9} & \textbf{15.3} \\
	 & q\sftsm & 48.3 & 52.7 & \textbf{15.3} \\
	\cmidrule(lr){1-2} \cmidrule(lr){3-4} \cmidrule(lr){5-5}
	\multirow{4}{*}{\rotatebox[origin=c]{90}{\textbf{Ll2-13b}}} & Original (4-bit) & 54.7 & 59.6 & 15.5 \\
	 & qLoRA & 55.0 & 60.7 & 18.2 \\
	 & q\sftrigl & \textbf{55.5} & 61.5 & \textbf{18.8} \\
	 & q\sftsm & 55.4 & \textbf{61.7} & 18.4 \\
	\bottomrule
\end{tabular}
    \vskip -0.1in
\end{table}

\section{Results}

\subsection{Main Results}
We present the main results of our experiments in \cref{tab:main-results}. For Llama2-7b, we find that \sftrigl outperforms LoRA and (IA)$^3$ consistently across evaluation benchmarks and instruction tuning datasets, including Flan v2, GPT4-Alpaca, and Tülu v2. For Llama2-13b, \sftrigl similarly outperforms all baselines when fine-tuned on Flan v2 and GPT4-Alpaca; however, we report more mixed results for Tülu v2.\footnote{It is possible that the hyperparameters chosen for \spiel during the hyperparameter search on Flan v2 do not always transfer well to Tülu v2, which is a much larger dataset with a different distribution of task types.} Nonetheless,  \sftrigl is superior in 5 out of the 6 combinations of scales and instruction tuning datasets. Hence, \sftrigl appears to be the strongest method overall. 

\textbf{AG vs MA} Comparing the two \spiel growth criteria, there appears to be a trade-off between performance and memory usage, with the more memory-efficient \sftsm generally performing a little worse than \sftrigl, except for a few cases, such as HumanEval evaluation for Llama2-7b. Since GPT4-Alpaca (used for HumanEval evaluation) has the same amount of training examples as Flan v2, we rule out that this difference is due to different levels of sample efficiency between \sftrigl and \sftsm.

\textbf{Ceiling} We find that \sftrigl generally matches the performance of full fine-tuning on most tasks, except GSM and HumanEval when fine-tuned on Tülu v2, where FullFT vastly outperforms all PEFT methods. This effect was also observed by \citet{ivison2023camels} in their qLoRA evaluation. We note that the maximum sequence length we use during fine-tuning is 2,048, whereas \citet{ivison2023camels} use 8,192 for full fine-tuning and 4,096 for qLoRA. It is possible that the shorter sequence length for PEFT has a significant effect on downstream performance for open-ended generation tasks, or that PEFT is weaker on these tasks in general, perhaps due to its inability to modify the input and output embedding layers. This warrants further investigation as part of future work.

\textbf{Quantization} According to the scores in Table~\ref{tab:quantized-results}, we find that 4-bit quantization results in only a modest reduction in performance across PEFT methods, and their relative performance remains similar. In fact, \sftrigl is superior again to both LoRA and (IA)$^3$ by an even higher margin compared to \cref{tab:main-results}. These results demonstrate that \spiel is a competitive PEFT method even in extremely resource-constrained scenarios.

\textbf{Compositionality} Finally, in \cref{tab:main-results} we also study whether \textit{multiple} PEFT adapters, trained on different data mixtures, can be composed with the LLM. We find that composing GPT4-Alpaca and Tülu v2 adapters increases the performance on HumanEval for PEFT, compared to the individual adapters. However \sftrigl performs favorably in composition compared to LoRA as it yields a larger boost for 7B (+1.7 vs +0.4) and an overall better performance on 13B.

\subsection{Training Speed and Memory Efficiency}
\begin{table}[t]

\caption{GPU memory requirements (in GB) and average time per training step (in seconds) for fine-tuning \sft and LoRA on Flan v2 on an A100 GPU. We report values either without (left) or with (right) activation checkpointing.}
    \label{tab:memory}
    \vskip 0.15in
    \def\arraytretch{0.9}
    \footnotesize
    \centering
    \begin{tabular}{lcccc}
        \toprule
         \textbf{Method} & \multicolumn{2}{c}{LlaMA 2 7b} & \multicolumn{2}{c}{LlaMA 2 13b}\\
          & Mem. $\downarrow$ & Time $\downarrow$ & Mem. $\downarrow$ & Time $\downarrow$ \\
         \midrule
         LoRA & 40 / 18 & \textbf{30.5} / 42.5 & 66 / \textbf{31} & \textbf{45.9} / \textbf{64.0} \\
         \sftrigl & 34 / 20 & 33.4 / 44.8 & 56 / 36 & 56.2 / 76.3 \\
         \sftsm & \textbf{30} / \textbf{17} & 30.6 / \textbf{41.7} & \textbf{51} / \textbf{31} & 50.6 / 70.9 \\
         
         \midrule
         qLoRA & 30 / \phantom{0}\textbf{8} & \textbf{38.5} / \textbf{55.2} & 46 / \textbf{12} & \textbf{63.6} / \phantom{0}\textbf{95.3} \\
         q\sftrigl & 26 / 13 & 42.8 / 58.4 & 40 / 19 & 73.4 / 101.1 \\
         q\sftsm & \textbf{22} / 10 & 39.6 / 55.5 & \textbf{35} / 14 & 70.1 / \phantom{0}97.3 \\
         
         \bottomrule
    \end{tabular}
\end{table}

In Table~\ref{tab:memory}, we compare the memory requirements and training time of LoRA, \spiel and their quantized variants. We consider two settings, with and without activation checkpointing.\footnote{Note that ``activation checkpointing'' is a synonym of ``gradient checkpointing''.} We define the memory requirements to be the minimum amount of GPU memory needed to complete training successfully. We provide more details on our measurements in \cref{app:memory_reqs}.

We find that LoRA, \sftrigl and \sftsm are broadly similar in terms of speed and memory requirements. \sftsm is consistently somewhat faster and more memory efficient than \sftrigl. Without activation checkpointing, we find that the \spiel methods are more memory efficient than LoRA; however, activation checkpointing is especially effective in reducing LoRA's memory consumption, likely because LoRA stores more intermediate activations due to the parallel computation of the LoRA update. When both quantization and activation checkpointing are applied, LoRA is the most memory-efficient technique by a small margin.\footnote{We note that, unlike our q\spiel implementation, qLoRA employs a paged optimizer, which might explain why there is a larger memory reduction for qLoRA. While q\spiel is also compatible with the use of paged optimizers, we leave the actual implementation for future work.} We also find that LoRA is generally slightly faster than the \spiel methods, despite being generally outperformed by \sftrigl, especially with quantization.

As expected, quantization and activation checkpointing both trade a slowdown in training speed for a reduction in memory usage. However, as a more general finding, these results would suggest that activation checkpointing should be prioritized ahead of quantization in most cases when performing PEFT on LLMs of this size ($\geq$ 7B), as a larger memory saving is achieved for an only marginally larger slowdown, and without incurring any cost in performance.

\subsection{Parameter Ages}
\begin{figure}[t]
    \centering
    \includegraphics[width=0.93\columnwidth]{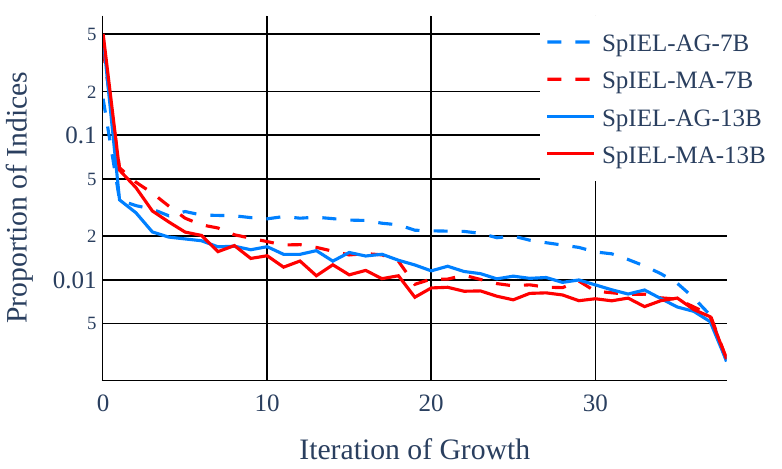}
    \caption{Proportion of indices with a certain age (i.e., the iteration when they were last grown) of the converged $\veta$ after training on the Flan v2 dataset.}
    \label{fig:parameter_ages}
\end{figure}

To shed light on the training dynamics of \spiel, we study the age of each index (i.e., the iteration when it was last grown) of the converged parameters. We plot the proportion of indices grown at a certain iteration in
\cref{fig:parameter_ages}, based on models trained on Flan v2. In general, \sftsm introduces fewer new parameter indices later in training compared to \sftrigl. We also find that 13b models tend to retain parameter indices found earlier in training compared to their 7b counterparts. This preference for earlier index sets is not necessarily desirable as it might reflect the fact that \sftsm and 13b models tend to get stuck in early local minima. We speculate that better schedules for dropping and growth might alleviate this issue in the future.

\section{Conclusions and Future Work}
We have proposed \spiel, the first method (to our knowledge) for Sparse Fine-Tuning (\sft) that is not only parameter-efficient but also memory-efficient. This allows it to scale to Large Language Models. Taking inspiration from iterative methods for sparse pre-training, we alternate among phases where we update, drop, and then grow parameters. In particular, we introduce two new criteria for parameter growth: Accumulated Gradients across multiple iterations (\sftrigl) and Momentum Approximation (\sftsm), where we reuse information tracked by optimizers like SM3. We find that \spiel often surpasses alternative PEFT methods such as (IA)$^3$ and LoRA in terms of performance and is comparable to them in terms of memory and time efficiency.

We hope that these promising results will encourage further research into sparse fine-tuning. Future work may include extending \spiel to the embedding layers of LLMs; improving the efficiency of the backward pass for a linear layer (see \cref{app:sft-backward}); considering more advanced dropping/growth criteria, especially those which would enable adaptive redistribution of tunable parameters across parameters tensors. This would stand in contrast to the current setup where the number of tunable parameters in each tensor is fixed.

\section*{Acknowledgments}
Alan Ansell wishes to thank David and Claudia Harding for their generous support via the Harding Distinguished Postgraduate Scholarship Programme. This work has been in part supported by the UK Research and Innovation (UKRI) Frontier Research Grant EP/Y031350/1 (the UK government’s funding guarantee for ERC Advanced Grants) awarded to Anna Korhonen. Ivan Vuli\'{c} has also been supported by a personal Royal Society University Research Fellowship \textit{`Inclusive and Sustainable Language Technology for a Truly Multilingual World'} (no 221137; 2022-). Hannah Sterz thanks the Cambridge Trust for their support via the International Scholarship.

\bibliography{anthology,custom}
\bibliographystyle{icml2024}

\appendix

\section{\sft vs LoRA Visualization}
\begin{figure}[th]
    \centering
    \includegraphics[width=0.88\columnwidth]{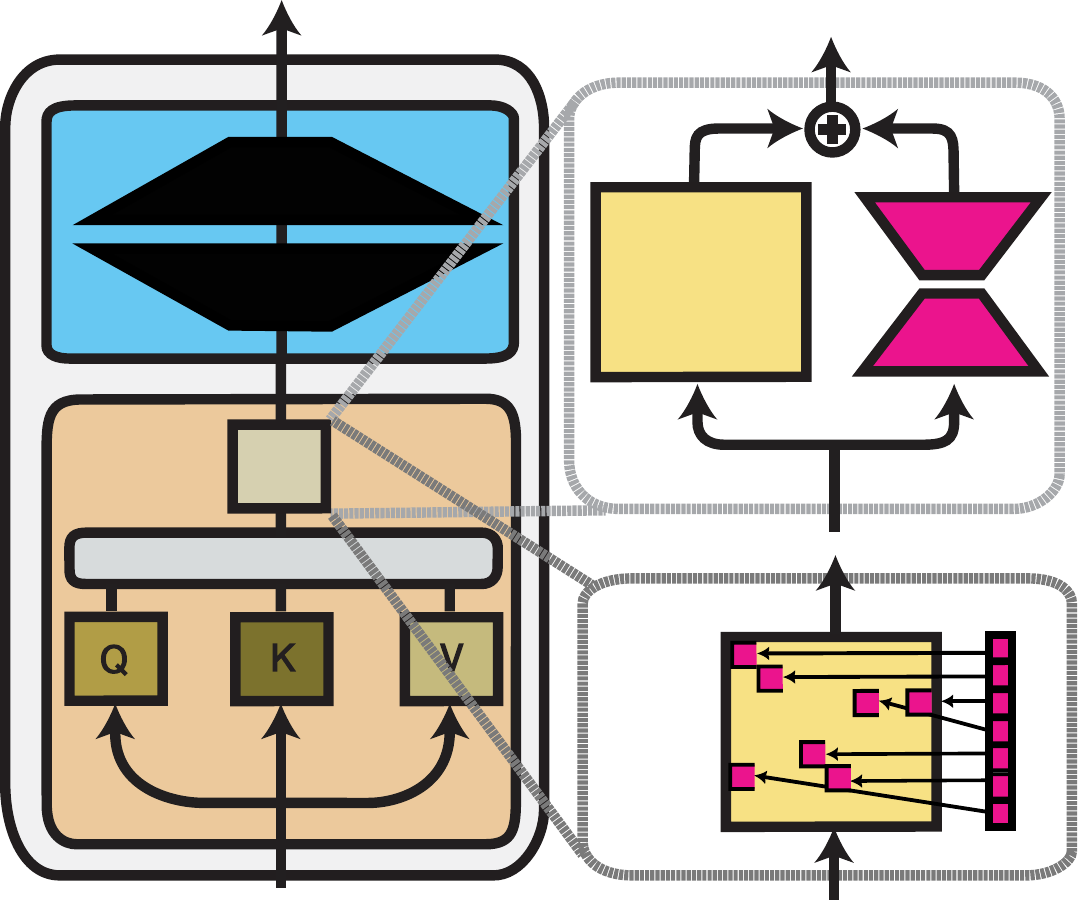}
    \vspace{-1mm}
    \caption{\sft (lower right) versus LoRA (upper right) applied to a linear layer (the output projection of self-attention) of a Transformer block.}
    \label{fig:peft}
    \vspace{-1mm}
\end{figure}

\section{Evaluation Setup}
\label{app:eval_setup}
Following \citet{wang2023far} and \citet{ivison2023camels}, we evaluate the instruction-tuned LLMs on a range of benchmarks to capture the following abilities:

\noindent
\textbf{Factuality}: Massively Multitask Language Understanding \citep[MMLU;][]{hendrycks2021measuring} requires the model to pick an answer from 4 candidates and covers 57 subjects including STEM, humanities, social sciences, and other disciplines. We evaluate models in a 5-shot setting and report their accuracy.

\noindent
\textbf{Reasoning}: We sub-sample 40 examples per task from BIG-Bench Hard \citep[BBH;][]{suzgun2022challenging}, a suite of the 23 most challenging tasks from BIG-Bench. We also randomly select a subset of 200 examples from Grade School Math \citep[GSM;][]{cobbe2021gsm8k}, a collection of math problems in linguistic form. Both benchmarks require open-ended generation. We evaluate models in a 3-shot setting with BBH and 8-shot setting with GSM, and report their exact match (EM).%

\noindent
\textbf{Multilinguality}: We choose 100 examples per language from Typologically Diverse Question Answering \citep[TyDiQA;][]{clark2020tydiqa}, a dataset for extractive question answering in 11 languages. We follow the Gold Standard setup where each document is guaranteed to contain the answer span. We evaluate models in a 1-shot setting and report F1.

\noindent
\textbf{Coding}: HumanEval \citep{chen2021evaluating} is a dataset for synthesizing programs from docstrings. We evaluate models with a temperature of 0.1 and report their precision at 1 (P@1)\footnote{Here we differ from \citet{wang2023far} in order to reduce the variance across evaluation runs.}.

\section{Training Details and Hyperparameters}
\label{app:training_details}

To select the most important hyperparameters of the PEFT methods, we perform a hyperparameter search as detailed in Appendix~\ref{app:hpsearch} with respect to (i) the number of trainable parameters (determined by rank $r$ in LoRA or corresponding density in \spiel)\footnote{
The number of trainable parameters for (IA)$^3$ is not tunable.}; (ii) the learning rate; (iii) weight decay strength $\lambda$ for \spiel methods. We use the found hyperparameters of each PEFT method for all fine-tuning experiments.

For \spiel, we update the set of trainable parameters $\veta$ every $S = 20$ steps, fix the initial update rate $\xi$ to 0.2, and for \sftrigl, we set the length $\gamma$ of the gradient estimation phase to 5 steps. Note that we have performed only minimal manual tuning on the values of $S$, $\xi$ and $\gamma$ due to a large number of large-scale experiments coupled with constraints on our computational budget; it is possible that other values would yield better results.

We follow \citet{wang2023far}'s choice for the remaining hyperparameters: we always train for two epochs, and apply linear learning rate decay following warmup over the first 3\% of training steps. The batch size is fixed at 128 and the maximum sequence length at 2,048, with longer sequences truncated. The LoRA dropout rate is set to 0.1 and LoRA $\alpha$ to 16. LoRA and \spiel are applied to all linear Transformer block layers. The pretrained parameters are stored in \texttt{bfloat16} data type, while the PEFT parameters are stored in \texttt{float32}, except for quantized training, where they are also stored in \texttt{bfloat16}.

\section{Hyperparameter Search}
\label{app:hpsearch}
We first performed a grid search over (i) learning rate and (ii) number of PEFT parameters (determined by rank $r$ for LoRA and $d_\vphi$ for \spiel), except for (IA)$^3$ where this number is fixed. For the \spiel methods, we also wished to establish a good value for (iii) weight decay strength $\lambda$. Since we generally obtained a good performance with the highest tested equivalent rank of $r = 64$, and being the most highly parameterized setting this was the most likely to benefit from regularization, we searched over a range of $\lambda$ values with $r$ set to 64 and the learning rate to the best value found in the initial search.

We searched for the optimal learning rate in the range of $\{\num{3e-6}, \num{1e-5}, \num{3e-5}, \num{1e-4}\}$ for LoRA and \sftrigl, $\{\num{4e-4}, \num{7e-4}, \num{1e-3}, \num{2e-3}\}$ for \sftsm (the SM3 optimizer generally benefits from higher learning rates than Adam), and $\{\num{1e-5}, \num{3e-5}, \num{1e-4}, \num{3e-4}\}$ for (IA)$^3$. %

As for the number of PEFT parameters, we searched over the range $r = \{8, 16, 32, 64\}$ for LoRA and the equivalent $d_\vphi$ for the \spiel methods. For LLaMA2-7b, these values correspond to 0.30\%, 0.59\%, 1.2\% and 2.3\% of the total parameter count respectively, and for LLaMA2-13b, they correspond to 0.24\%, 0.48\%, 0.95\% and 1.9\%.

We searched over $\lambda$ values in the range $\{0, 1, 3, 10, 30\}$ for \sftrigl and $\{0, 0.1, 0.3, 1, 3\}$ for \sftsm (since we use higher learning rates for \sftsm, a lower weight decay strength is required to have the same effect).

Each hyperparameter setting was evaluated by training on the Flan v2 subset and taking the 5-shot performance on the MMLU development set.

\begin{figure*}
    \centering
    \begin{subfigure}{0.3\linewidth}
        \includegraphics[width=\linewidth]{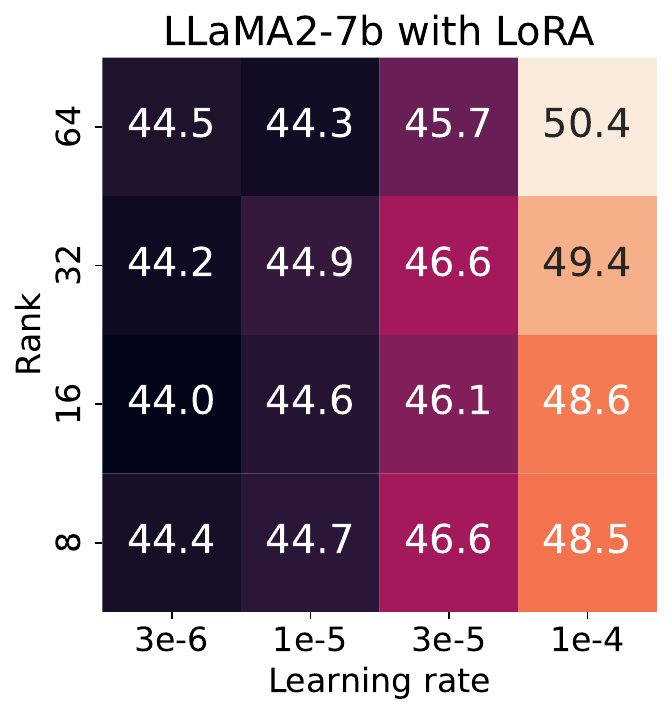}
    \end{subfigure}
    \begin{subfigure}{0.3\linewidth}
        \includegraphics[width=\linewidth]{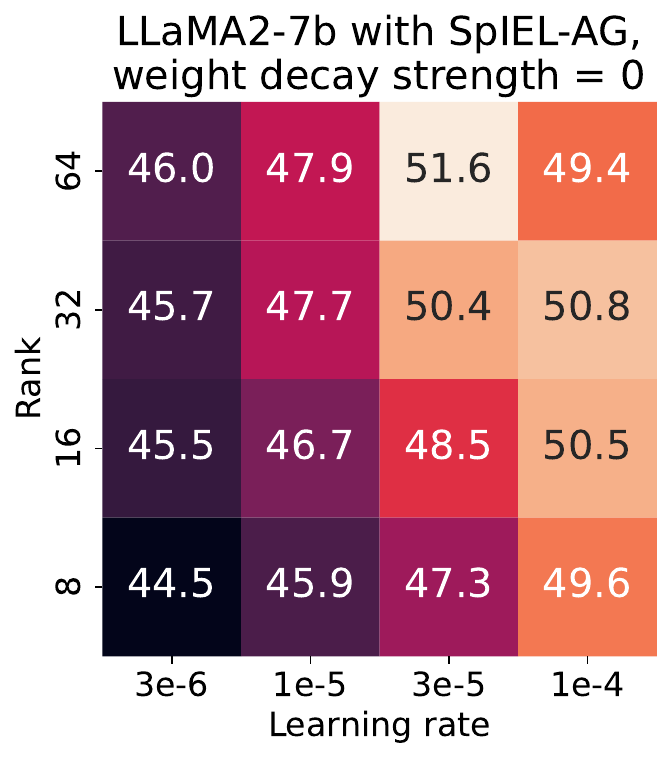}
    \end{subfigure}
    \begin{subfigure}{0.3\linewidth}
        \includegraphics[width=\linewidth]{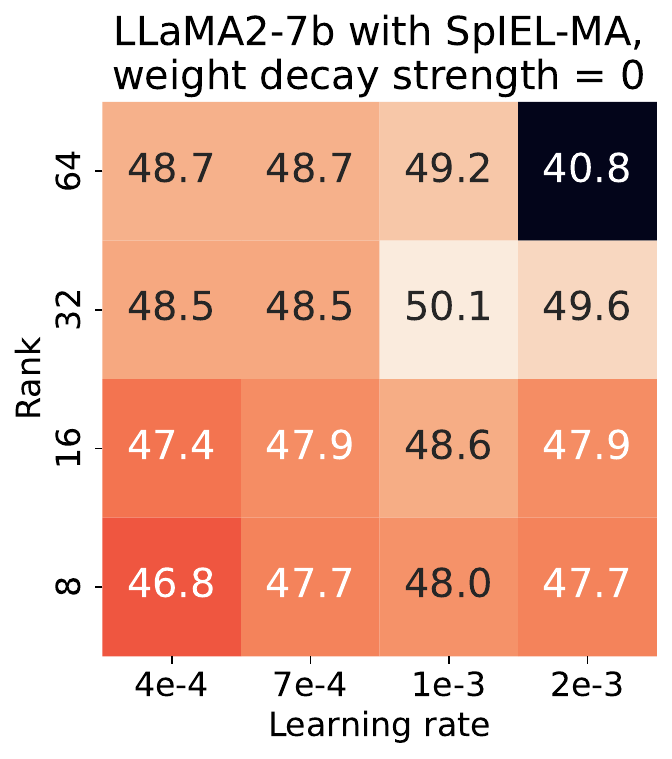}
    \end{subfigure}
    \begin{subfigure}{0.3\linewidth}
        \includegraphics[width=\linewidth]{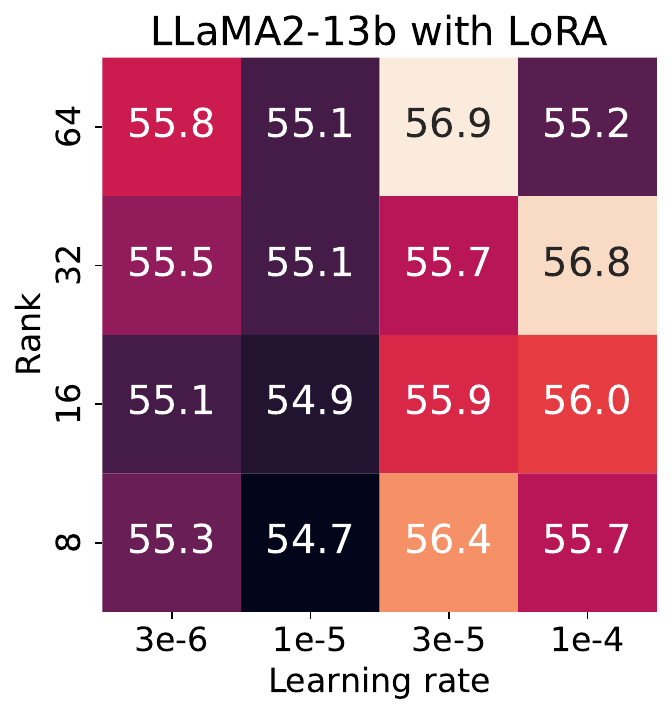}
    \end{subfigure}
    \begin{subfigure}{0.3\linewidth}
        \includegraphics[width=\linewidth]{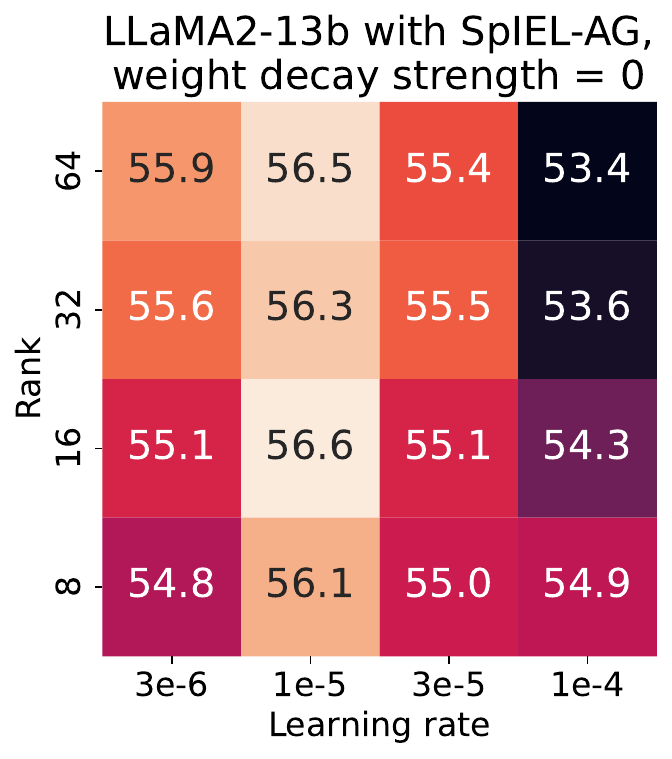}
    \end{subfigure}
    \begin{subfigure}{0.3\linewidth}
        \includegraphics[width=\linewidth]{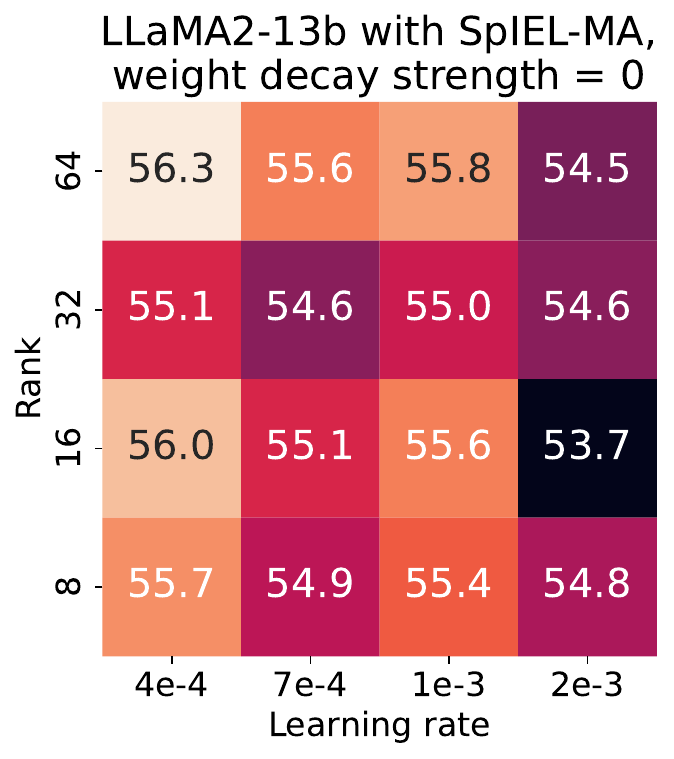}
    \end{subfigure}
    \begin{subfigure}{0.3\linewidth}
        \includegraphics[width=\linewidth]{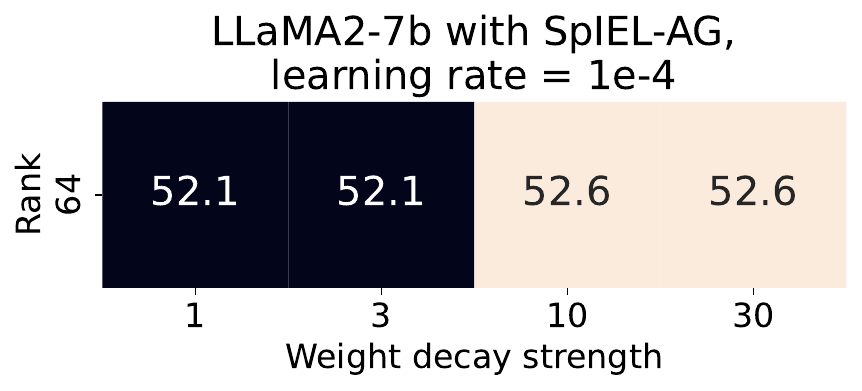}
    \end{subfigure}
    \begin{subfigure}{0.3\linewidth}
        \includegraphics[width=\linewidth]{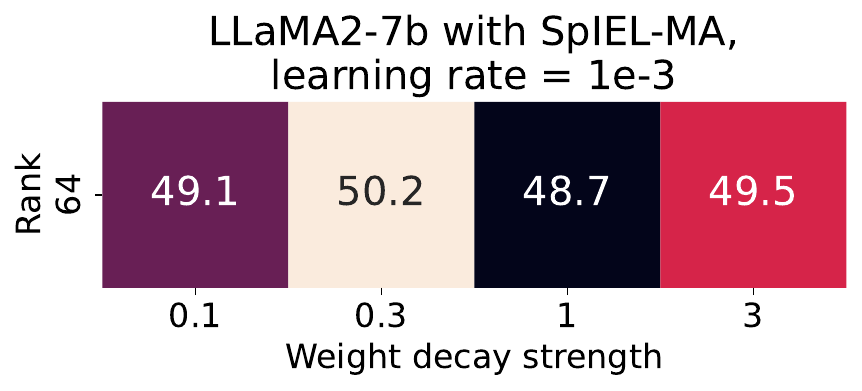}
    \end{subfigure}
     \begin{subfigure}{0.3\linewidth}
        \includegraphics[width=\linewidth]{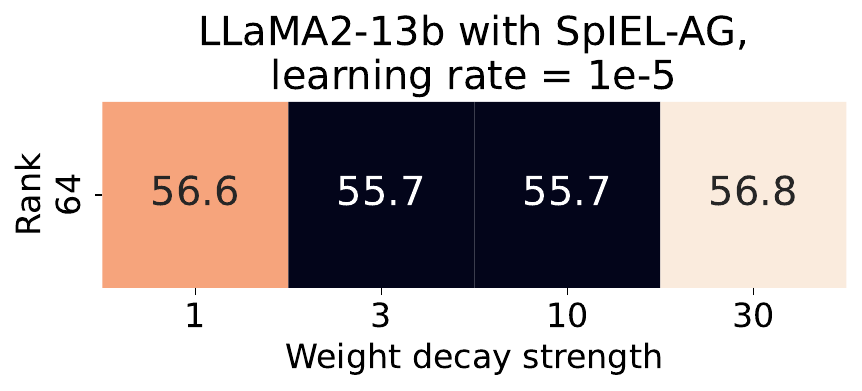}
    \end{subfigure}
    \begin{subfigure}{0.3\linewidth}
        \includegraphics[width=\linewidth]{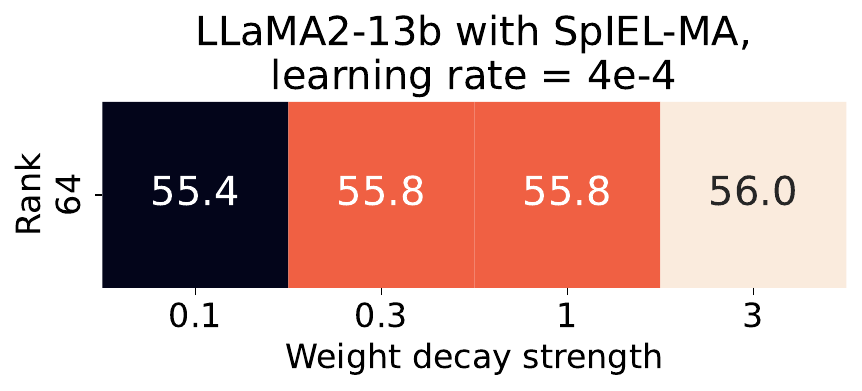}
    \end{subfigure}
    \caption{Hyperparameter search results.}
    \label{fig:hp-search}
\end{figure*}

We present the full results of the hyperparameter search in Figure \ref{fig:hp-search} and summarize the best values found for each configuration in Table \ref{tab:hp-search-results}.

\begin{table}[ht]
    \centering
    \resizebox{\columnwidth}{!}{
    \begin{tabular}{lcccccc}
         \toprule
         & \multicolumn{3}{c}{LLaMA2-7b} & \multicolumn{3}{c}{LLaMA2-13b}\\
         \cmidrule(lr){2-4} \cmidrule(lr){5-7}
         \textbf{Method} & lr & $r$ & $\lambda$ & lr & $r$ & $\lambda$ \\
         \cmidrule(lr){1-1} \cmidrule(lr){2-4} \cmidrule(lr){5-7}
         (IA)$^3$ & $\num{3e-4}$ & - & - & $\num{1e-4}$ & - & - \\
         LoRA & $\num{1e-4}$ & 64 & - & $\num{3e-5}$ & 64 & - \\
         \sftrigl & $\num{1e-4}$ & 64 & 30 & $\num{1e-5}$ & 64 & 30 \\
         \sftsm & $\num{1e-3}$ & 64 & 0.3 & $\num{4e-4}$ & 64 & 0 \\
         \hline
         Full FT & $\num{2e-5}$ & - & - & $\num{2e-5}$ & - & - \\
         \bottomrule
    \end{tabular}
    }
    \caption{Optimal settings yielded by hyperparameter search for each PEFT method and model size.}
    \label{tab:hp-search-results}
\end{table}

After our main experiments, we performed some additional exploration of the drop/growth schedule, adjusting both the frequency $\gamma$ of dropping/growth and the schedule $k(i, t)$ of number of parameters to drop/grow at each update. We considered drop/growth frequency in the range $\{10, 20, 40, 80\}$ steps and both our default linear schedule and the cosine schedule of \citet{evci-etal-2020-rigging}, where
\begin{align}
    k(i, t) = \frac{\xi}{2}\Big(1 + \cos\Big(\frac{t\pi}{T}\Big)\Big) d_{\vphi^{(i)}},
\end{align}
recalling that $\xi$ is the initial update rate and $T$ is the total number of training steps. Table \ref{tab:schedule-search} presents our results for \sftrigl when fine-tuning LLaMA2-7b on the Flan v2 subset with the other hyperparameters as shown in Table \ref{tab:hp-search-results}. We find that there is not a clear preference for a linear or cosine schedule, but that a higher update frequency of 10 might be better than the 20 steps we used in our main experiments.

\begin{table}[ht]
    \centering
    \footnotesize
    \begin{tabular}{lccc}
         \toprule
         \multicolumn{2}{c}{Schedule} & \multicolumn{2}{c}{Benchmark} \\
         \cmidrule(lr){1-2} \cmidrule(lr){3-4}
         \textbf{Type} & \textbf{Steps} & \textbf{MMLU} & \textbf{TyDiQA} \\
         \cmidrule(lr){1-1} \cmidrule(lr){2-2} \cmidrule(lr){3-3} \cmidrule(lr){4-4}
         \multirow{4}{*}{Linear} & 10 & \textbf{52.0} & 56.4 \\
         & 20 & 50.7 & 56.2 \\
         & 40 & 51.5 & \textbf{56.5} \\
         & 80 & 51.3 & 56.1 \\
         \cmidrule(lr){1-1} \cmidrule(lr){2-2} \cmidrule(lr){3-3} \cmidrule(lr){4-4}
         \multirow{4}{*}{Cosine} & 10 & \textbf{51.8} & 56.3 \\
         & 20 & 51.6 & \textbf{56.5} \\
         & 40 & 50.2 & \textbf{56.5} \\
         & 80 & 50.9 & 55.4 \\
         \bottomrule
    \end{tabular}
    \caption{\sftrigl results when fine-tuning LLaMA2-7b on Flan v2 subset with various frequencies and update schedule types.}
    \label{tab:schedule-search}
\end{table}

For full fine-tuning, we use the learning rate of $\num{2e-5}$ from \citet{wang2023far}. To fit the model into a single A100, we additionally resort to activation checkpointing and paged optimization.

\section{Analysis of the Backward Pass for \spiel}
\label{app:sft-backward}
Consider the forward pass when sparsely fine-tuning a linear layer. We have:
\begin{align}
    Y = X(W + \Delta),
\end{align}
where the input $X \in \sR^{b \times d_\text{in}}$, the pretrained weight matrix $W$ and its sparse delta $\Delta \in \sR^{d_\text{in} \times d_\text{out}}$, and the output $Y \in \sR^{b \times d_\text{out}}$, with $b$, $d_\text{in}$ and $d_\text{out}$ being the batch size and input and output dimensions respectively. It can be shown that
\begin{align} \label{eq:naive-sft-backward}
    \frac{\partial \gL}{\partial \Delta} = X^\top \frac{\partial \gL}{\partial Y}.
\end{align}
However, we only need the entries of $\frac{\partial \gL}{\partial \Delta}$ corresponding to the currently active indices of $\Delta$. Let $\bm{r} \in \{1, 2, ..., h\}^N, \bm{c} \in \{1, 2, ..., w\}^N$, where $h$ and $w$ are the height and width of $\Delta$ respectively, denote the $N$ active indices of $\Delta$, i.e. $(r_i, c_i)$ denotes the position of the $i$th active index in $\Delta$. Then, if we define $g_i$ to be the gradient of the $i$-th active index of $\Delta$, we have
\begin{align}
    g_i &= \frac{\partial \gL}{\partial \Delta_{r_i, c_i}} \\
        &= \Big(X^\top \frac{\partial \gL}{\partial Y}\Big)_{r_i, c_i} \\
        &= X_{:, r_i}^\top \Big(\frac{\partial \gL}{\partial Y}\Big)_{:, c_i}.
\end{align}
That is, the gradient of the $i$-th sparse update to $W$ is given by the dot product of the $r_i$-th column of $X$ and the $c_i$-th column of the gradient of the output $Y$. We can compute the gradient of the sparse updates in this manner with $bN$ FLOPs, which is an enormous theoretical improvement over the $b d_\text{in} d_\text{out}$ FLOPs required to naively perform the full matrix multiplication in (\ref{eq:naive-sft-backward}) and gather the relevant indices from the result, as the \sft density $\frac{N}{d_\text{in} d_\text{out}} \ll 1$.

While calculating the sparse delta gradient in this manner entails a great reduction in FLOPs required, it is not so easy to exploit this reduction effectively on a GPU to speed the operation up. Writing an efficient CUDA kernel for this operation is ongoing work, and the speed results presented in this paper were obtained using the naive ``gather from the full matrix product'' method.%

\section{Measurement of Memory and Time Requirements}
\label{app:memory_reqs}
To measure the memory requirements of PEFT methods, we use PyTorch's \texttt{set\_per\_process\_memory\_fraction} function to limit the total available GPU memory, and perform a binary search at 1 GiB granularity to find the lowest limit at which training can run successfully. For each limit we test, we run 30 steps of Flan v2 training with (equivalent) LoRA rank 64. We measure the time as the mean duration of each of these 30 steps for the lowest passing memory limit. All our experiments are run on a single A100 GPU.

We note that this may differ from the peak memory usage during (unconstrained) training, since deep learning frameworks such as PyTorch may allocate more memory than they actually require for efficiency reasons.

\end{document}